\documentclass[journal]{IEEEtran}
\usepackage[numbers,sort&compress]{natbib}
\usepackage{tabu}
\usepackage{graphicx}
\usepackage{algorithmic}
\usepackage{algorithm}
\usepackage{multirow}
\usepackage{multicol}
\usepackage{amsmath}
\usepackage{amssymb}
\ifCLASSINFOpdf
\else
\fi
\begin{document}
\title{A Unified Joint Maximum Mean Discrepancy for Domain Adaptation}

\author{Wei~Wang,
        Baopu~Li,
        Shuhui~Yang,
        %Mengzhu~Wang,
        Jing~Sun,
        Zhengming~Ding,
        Junyang~Chen,
        Xiao~Dong,
        Zhihui~Wang,
        and~Haojie~Li$^{\star}$% <-this % stops a space
\thanks{W. Wang, S. Yang, J. Sun, Z. Wang and H. Li are with the DUT-RU International School of Information Science \& Engineering, Dalian University of Technology, Dalian, Liaoning, 116000, P.R. China email: WWLoveTransfer@mail.dlut.edu.cn, shu\_hui\_yang@163.com, sunjing616@mail.dlut.edu.cn, zhwang@dlut.edu.cn, hjli@dlut.edu.cn.}% <-this % stops a space
\thanks{B. Li is with Baidu Research, Sunnyvale CA 94089, USA email: bpli.cuhk@gmail.com.}% <-this % stops a space
\thanks{Z. Ding is with Department of Computer Science, Tulane University, New Orleans LA 70118, USA email: zding1@tulane.edu.}
\thanks{J. Chen is with the College of Computer Science and Software Engineering, University of Shenzhen, Shenzhen, Guangdong, P.R. China email: yb77403@umac.mo.}
\thanks{X. Dong is with the School of Information Technology and Electrical Engineering (ITEE), University of Queensland, Brisbane, 4072, Australia email: dx.icandoit@gmail.com.}}


\markboth{IEEE Transactions on ***}%
{Shell \MakeLowercase{\textit{et al.}}: Bare Demo of IEEEtran.cls for IEEE Journals}



\maketitle

\begin{abstract}
Domain adaptation has received a lot of attention in recent years, and many algorithms have been proposed with impressive progress. However, it is still not fully explored concerning the joint probability distribution ($P(\textbf{\textit{X}},\textbf{\textit{Y}})$) distance for this problem, since its empirical estimation derived from the maximum mean discrepancy (joint maximum mean discrepancy, JMMD) will involve complex tensor-product operator that is hard to manipulate. To solve this issue, this paper theoretically derives a unified form of JMMD that is easy to optimize, and proves that the marginal, class conditional and weighted class conditional probability distribution distances are our special cases with different label kernels, among which the weighted class conditional one not only can realize feature alignment across domains in the category level, but also deal with imbalance dataset using the class prior probabilities. From the revealed unified JMMD, we illustrate that JMMD degrades the feature-label dependence (discriminability) that benefits to classification, and it is sensitive to the label distribution shift when the label kernel is the weighted class conditional one. Therefore, we leverage Hilbert Schmidt independence criterion and propose a novel MMD matrix to promote the dependence, and devise a novel label kernel that is robust to label distribution shift. Finally, we conduct extensive experiments on several cross-domain datasets to demonstrate the validity and effectiveness of the revealed theoretical results.
\end{abstract}


% \fi
\IEEEpeerreviewmaketitle

\section{Introduction}
\IEEEPARstart{A}{long} with massive appearances of new data, it is impossible to manually annotate them all at a labor-intensive and time-consuming expense, especially for on-line and real-time systems%~\cite{ouyang2,ouyang3,ouyang4}
. A plausible strategy is to reuse the model trained on existing dataset to the newly-coming one, while it often fails to achieve promising performance due to their probability distribution difference%~\cite{ouyang1}
. This urgently requires us to devise a versatile algorithm to handle such a great challenge. Fortunately,  domain adaptation (DA) as an effective technology
has attracted considerable attention, which allows for training and test datasets to follow divergent probability distributions, and commits to exploiting common knowledge across them by minimizing their probability distribution discrepancy desirably~\cite{CVPR1,CVPR2,CVPR3}.

\begin{figure}[t]
	\begin{center}
		\includegraphics[width=0.9\linewidth]{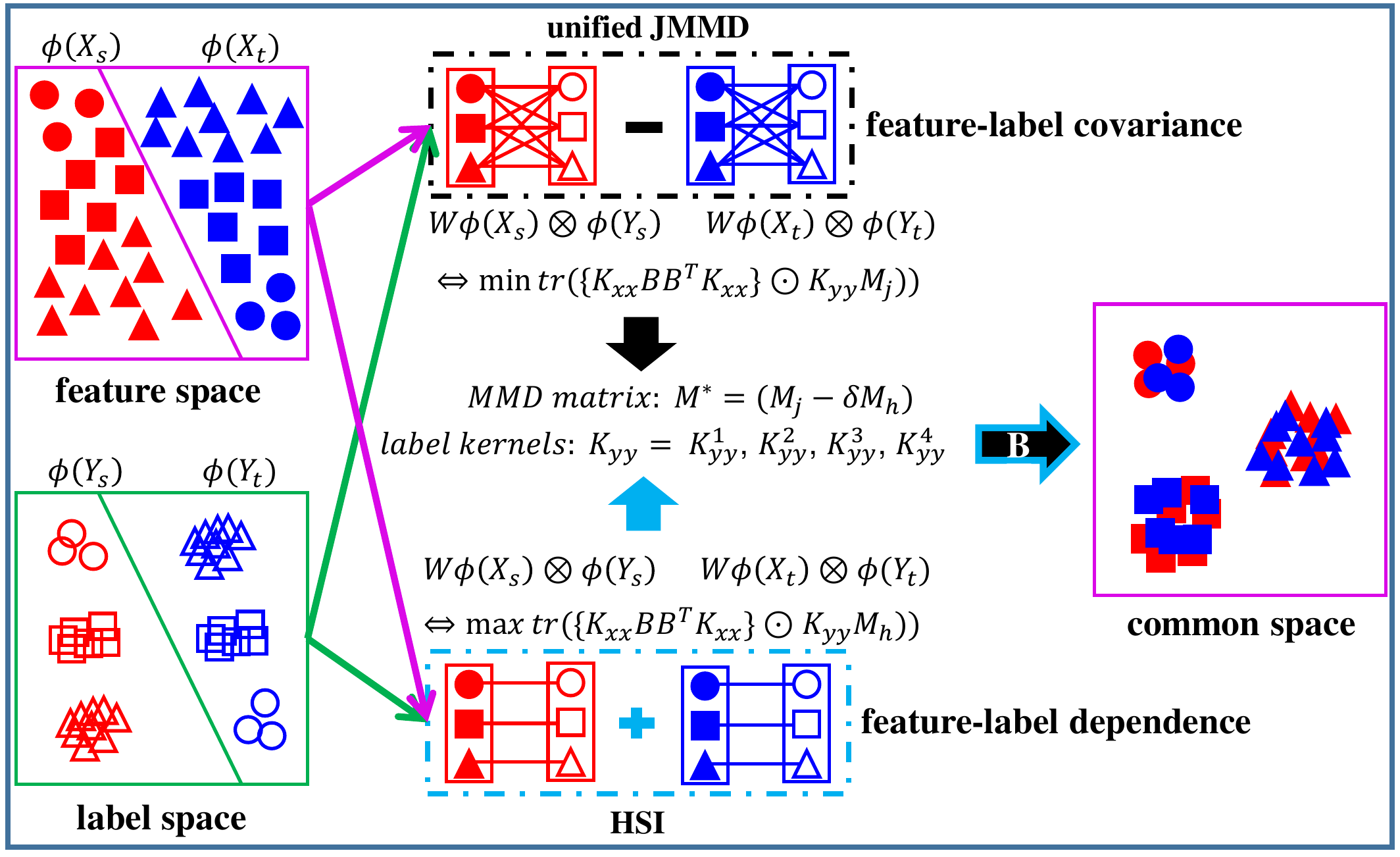}
	\end{center}
	\caption{The overview of proposed unified JMMD. The red and blue colors are used for the source and target domains, different shapes denote different categories, and the solid and hollow dots represent the feature and label representations of a data instance. The unified JMMD with a novel MMD matrix $\textbf{\textit{M}}^{*}$ and a novel label kernel $\textbf{\textit{K}}_{\textbf{\textit{y}}\textbf{\textit{y}}}^{4}$ is proposed, respectively ($\textbf{\textit{K}}_{\textbf{\textit{y}}\textbf{\textit{y}}}^{1}$, $\textbf{\textit{K}}_{\textbf{\textit{y}}\textbf{\textit{y}}}^{2}$, $\textbf{\textit{K}}_{\textbf{\textit{y}}\textbf{\textit{y}}}^{3}$ are three special cases of ours). $\textbf{\textit{B}}$ represents a projection matrix for a common feature space. $-$ in the module of unified JMMD means the correlation difference between source and target domains, and $+$ in the module of HSI means separately considering the dependence in both source domain and target domain.
		%Please refer to the section of introduction for details.
	}
	\label{fig1}
\end{figure}

\par A decisive problem in DA is how to formulate a favorable probability distribution distance that can be applied to measure the proximity of two different probability distributions, thus numerous probability distribution distance-metrics have been proposed over the years. For example, the Quadratic~\cite{BD} and Kullback-Leibler~\cite{KL} distances that are derived from the Bregman divergence and generated by different convex functions were introduced to explicitly match two different probability distributions (\textit{i.e.}, $P(\textbf{\textit{X}})$). However, it is inflexible to extend them to different models since they are parametric and require an intermediate density estimate~\cite{TCA}, and cannot describe more complicated probability distributions such as conditional (\textit{i.e.}, $P(\textbf{\textit{Y}}|\textbf{\textit{X}})$) and joint (\textit{i.e.}, $P(\textbf{\textit{X}},\textbf{\textit{Y}})$) probability distributions. 

\par The Wasserstein distance derived from optimal transport problem exploited a transportation plan to align two different marginal~\cite{OT1,OT2} or joint~\cite{OT3,OT4} probability distributions, but it is still not convenient to be applied to the traditional DA methods because it often comes down to a complex bi-level optimization problem~\cite{WDA}. Noticeably, the maximum mean discrepancy (MMD)~\cite{MMD}, a metric based on embedding of probability distribution in a reproducible kernel Hilbert space (RKHS), has been applied successfully in a wide range of problems thanks to its simplicity and solid theoretical foundation, such as transfer learning~\cite{TCA}, kernel Bayesian inference~\cite{KBP}, approximate Bayesian computation~\cite{KABC}, two-sample~\cite{MMD}, goodness-of-fit testing~\cite{GFT}, MMD GANs~\cite{GMMN} and auto-encoders~\cite{IMVA}, \textit{etc}.  

\par Although the MMD %approach 
has been successfully applied to establish the marginal~\cite{TCA,DAN}, class conditional~\cite{JDA,CAN} (\textit{i.e.}, $P(\textbf{\textit{X}}|\textbf{\textit{Y}})$) and weighted class conditional~\cite{BDA} (\textit{i.e.}, $P(\textbf{\textit{Y}})P(\textbf{\textit{X}}|\textbf{\textit{Y}})$) probability distribution distances, it is still not fully explored so far concerning the joint probability distribution distance, especially for the traditional DA methods,  since its empirical estimation derived from the MMD (joint maximum mean discrepancy, JMMD) involves complex tensor-product operator. % that is nontrivial to manipulate~\cite{JAN}. 
In addition, the marginal probability distribution distance aligns the source and target feature distributions holistically but ignores prior label information. To realize more effective knowledge transfer, the class conditional one matches their distributions in a category level while it does not consider the problem of imbalanced dataset. To this end, the weighted class conditional one imposes bigger weights on the larger categories by $P(\textbf{\textit{Y}})$ (\textit{i.e.}, $P(\textbf{\textit{Y}}_s)P(\textbf{\textit{X}}_s)-P(\textbf{\textit{Y}}_t)P(\textbf{\textit{X}}_t)$) based on the assumption that $P(\textbf{\textit{Y}}_s)=P(\textbf{\textit{Y}}_t)$. However, their feature distributions are misaligned when $P(\textbf{\textit{Y}}_s)\neq P(\textbf{\textit{Y}}_t)$. Moreover, recent works indicate that the procedure of feature distribution alignment degrades its discriminability~\cite{ALN,BSP} but lack of theoretical supports. 

\par 
%To this end, this paper overcomes the problem of complicated tensor-product operator to theoretically derive a unified form of JMMD that is easy to optimize. 
%Moreover, we prove that the marginal, class conditional and weighted class conditional probability distribution distances are our special cases with different label kernels. Specifically, as can be seen from Figure.\ref{fig1},
To solve the above problems, we first project the data features and their corresponding labels to a reproducible kernel Hilbert space that facilitates construction of the JMMD. And JMMD aims to build a tensor-product that can describe the relationship between source domain and target domain, and gradually promotes the coincidence of the distributions between two domains. Two issues remain to be overcome in this process, that is, the complexity involved in the computational process of tensor-product and the reduction of the dependence between features and labels. To mitigate the above two problems, we derive a unified form of JMMD that can avoid the tensor-product,  and prove that the marginal, class conditional and weighted class conditional probability distribution distances are our special cases with different label kernels. Based on this unified form, 
%? we have two new discoveries ? Then, 
a novel MMD matrix can be obtained by reducing  the MMD matrix for Hilbert Schmidt independence (HSI)~\cite{HSI}  (\textit{i.e.}, $\textbf{\textit{M}}_h$) from  the MMD matrix for JMMD (\textit{i.e.}, $\textbf{\textit{M}}_j$). By minimizing this novel MMD matrix, 
%and maximizing the feature-label dependence in HSI space 这一项已经包含在新的MMD矩阵了
we can further improve the feature discriminability, leading to a better DA capacity. Moreover, a new label kernel $\textbf{\textit{K}}_{\textbf{\textit{y}}\textbf{\textit{y}}}^4$ is devised to address the problem of label distribution shift.
%by maximizing the feature-label dependence. 
The whole pipeline of our algorithm is briefly depicted in  Fig.\ref{fig1}.

\par The main contributions of our work are summarized as below,
%\% do not mention the P(Y) problem at the beginning and place this point at the discussions

%\%

\begin{itemize}
	\item 
	We theoretically unify the previous three widely used  methods that include marginal, class conditional and weighted class probability distribution distances,
	yielding a better guidance to refine the JMMD for different problems in domain adaption.
	\item {A new MMD matrix is proposed by deduction of the MMD matrix for HSI from the one for JMMD  to improve the feature-label correlations.}
	\item {One new label reproducing kernel is suggested to deal with label distribution shift.} 
	\item {Extensive experiments are conducted on some benchmark datasets, validating the effectiveness of the proposed unified JMMD with the above two strategies.}
\end{itemize}

% 结合上面对图1的解释这里贡献点需要再讨论
%-------------------------------------------------------------------------
\section{Related Works}
\subsection{Maximum Mean Discrepancy} 
\par Gretton \textit{et al.} introduced a non-parametric distance estimate called the maximum mean discrepancy (MMD) that is designed by embedding distributions in a reproducible kernel Hilbert space (RKHS) to compute the distance between different marginal probability distributions~\cite{MMD}. Along this direction, Pan \textit{et al.} incorporated the MMD into a subspace learning method to learn some transfer components across the two domains~\cite{TCA}, Duan \textit{et al.} embedded the MMD into a multiple kernel learning framework, to jointly learn a kernel function and a robust classifier by minimizing the structural risk functional and the distribution mismatching~\cite{DTMKL}, Long \textit{et al.} down-weighted the source instances that are irrelevant to target ones to realize more effective knowledge transfer using the MMD~\cite{TJM}. Ghifary \textit{et al.} employed the MMD to deal with distribution bias in a simple neural network~\cite{DANN}, Tzeng \textit{et al.} proposed a domain confusion loss based on the MMD in a deep neural network~\cite{DDC} (AlexNet~\cite{AlexNet}), Long \textit{et al.} presented a deep adaptation network (DAN) where the multi-kernel MMD was applied to the hidden representations of all task-specific layers~\cite{DAN}. 

\par To further improve the representation ability of the MMD respecting more complicated probability distribution distances, Long \textit{et al.} resorted to the sufﬁcient statistics of class conditional probability distribution (\textit{i.e.}, $P(\textbf{\textit{X}}|\textbf{\textit{Y}})$) to approximate the conditional probability distribution (\textit{i.e.}, $P(\textbf{\textit{Y}}|\textbf{\textit{X}})$) distance, referred to as the class-wise MMD, which can be conveniently optimized in most subspace-learning approaches~\cite{VDA,JGSA,DICD,TIT,LPJT}. Kang \textit{et al.} raised a class-aware sampling strategy to establish the class-wise MMD that can be effectively estimated~\cite{CAN} in a deep neural network. 
To deal with imbalanced dataset, Wang \textit{et al.}~\cite{BDA} established a weighted class-wise MMD in which the class conditional probability distribution is imposed by the class prior probabilities (\textit{i.e.}, $P(\textbf{\textit{Y}})P(\textbf{\textit{X}}|\textbf{\textit{Y}})$). Concerning the label distribution shift across the source and target domains, Yan \textit{et al.}~\cite{WMMD} constructed a weighted MMD where the source data instances are multiplied by $P(\textbf{\textit{Y}}_t)/P(\textbf{\textit{Y}}_s)$. Moreover, Wang \textit{et al.} raised a dynamic balanced MMD to quantitatively account for the relative importance of marginal and conditional distributions~\cite{MEDA1,MEDA2}. Long \textit{et al.} constructed a transfer network by directly minimizing the joint probability distribution (\textit{i.e.}, $P(\textbf{\textit{X}},\textbf{\textit{Y}})$) distance of multiple domain-specific layers across the two domains using the joint maximum mean discrepancy (JMMD)~\cite{JAN}. However, the JMMD is hard to be applied to traditional DA methods since it involves complex tensor-product operator that is nontrivial to manipulate. This paper not only theoretically derives a unified format of the JMMD that is easy to optimize, but also proves that the MMD, class-wise MMD and weighted class-wise MMD are our special cases with different label kernels. %From the revealed unified JMMD, we provide  theoretical supports that feature alignment may damage its feature discriminability (feature-label dependence), and propose a novel MMD matrix to improve the dependence. Although the weighted class conditional probability distribution distance is the most effective one among them, it is sensitive to label distribution shift. Inspired by WMMD, we propose a novel label kernel to deal with this problem. Different from WMMD, we consider it in a category level to boost more effective knowledge transfer.
% 这里和边缘,条件,加权条件的区别是它们是我们的特殊情况.与它们不同的是,我们从理论上指出了它们破坏了特征和标签的依赖性,并且提供了解决方法. 此外,尽管加权条件是三者中的最好的,仍然存在对标签分布差异敏感的问题,借鉴WMMD的思想我们将其引入到JMMD中,提出了一种新的标签核. 与WMMD的区别是,它们是在边缘分布上整体考虑这个问题,我们是在JMMD这个统一的框架上在类别的层面考虑,因此能更有效的迁移,而且从理论上说明了为什么标签分布差异会导致错误迁移的问题.
\subsection{Hilbert Schmidt Independence}
\par The Hilbert Schmidt independence (HSI) that is formulated with Hilbert Schmidt norm in RKHS is used to measure the dependency between two random variables~\cite{HSI}. To preserve the important feature of data, Dorri \textit{et al.} regarded original and transformed data features as two random variants, and maximized their dependency during the probability distribution alignment~\cite{ACA,ACA2}. Yan \textit{et al.} minimized the dependency between the projected and domain features (\textit{e.g.} background information) to learn a domain-invariant subspace~\cite{MIDA}. Inspired by the classifier adaptation, Zhang \textit{et al.} proposed a projected HSI criterion to improve the feature-label correlation in an reconstruction DA framework~\cite{CSRTL}. Similar to the goal in~\cite{CSRTL}, we also aim to improve the feature-label correlation. However, they do not explore the relationship between the JMMD and HSI as we do in this work.  %since the JMMD is hard to manipulate, and
In addition, the HSI is computed  on the whole domains~\cite{CSRTL}. In contrast,  we separately compute the HSI on the source and target domains.
% reveal that the JMMD degrades feature-label dependency (HSI) and separately. Moreover, they utilize the centered variant of HSI but this paper unifies the JMMD and HSI with its uncentered variant to enable the formulation more neat with a novel MMD matrix.}

\section{Revisit Maximum Mean Discrepancy}

%\ZD{This section is too much. Since it is a revist of MMD, we can directly propose the weighted one in a unified model, and specify back to class-wise/domain-wise MMD. We do not need to split into so many sections. We can marge all algorithm related sections into one.}

\par We start with some notations and problem statement appeared in this paper.

\subsection{Notations and Problem Statement}

\par The bold-italic uppercase letter $\textbf{\textit{X}}$ denotes a matrix, %and the lowercase of $\textbf{\textit{x}}$ is a column vector.
$\textbf{\textit{X}}_{(ij)}$ is the value of $i$-th row and $j$-th column of $\textbf{\textit{X}}$, $\textbf{\textit{x}}_i$ is the $i$-th column vector of $\textbf{\textit{X}}$, and $\textbf{\textit{x}}_{(i)}$ is the $i$-th component of $\textbf{\textit{x}}$. The subscript $s$ (resp. $t$) represents the index of source (resp. target) domain. The superscripts $(c)$ and $\top$ denote the index of $c$-th category and transpose of a matrix. tr($\cdot$), $\langle\cdot,\cdot\rangle$, $\otimes$ and $\odot$ are the operators of trace, inner-product, tensor-product, and dot-product, respectively. $\textbf{I}$ and $\textbf{1}$ represent an identity matrix and a matrix whose elements are all 1. $\textbf{H}=\textbf{I}-\frac{1}{n}\textbf{1}_{n\times n}$ is the centering matrix. $||\cdot||_F$, $||\cdot||_2$, and $||\cdot||_{\mathcal{H}}$ denote $l_F$ norm, $l_2$ norm and Hilbert-Schmidt norm, respectively.

\noindent \textbf{Domain Adaptation.} Given a fully-labeled source domain $D_s=(\textbf{\textit{X}}_s, \textbf{\textit{Y}}_s)$ and an unlabeled target domain $D_t=(\textbf{\textit{X}}_t, \textbf{\textit{Y}}_t)$, where $\textbf{\textit{X}}_{s/t}\in \mathbf{R}^{m\times n_{s/t}}$, $\textbf{\textit{Y}}_{s/t}\in \mathbf{R}^{C\times n_{s/t}}$, $m$ is the feature dimensions, $n_{s/t}$ is the number of source or target data instances ($n_s+n_t=n_{st}$), and $C$ is the number of categories. Here $\textbf{\textit{Y}}_{s/t}$ is the soft labels, i.e., the probability of a data sample $\textbf{\textit{x}}_i$ that belongs to class $c$ is $\textbf{\textit{Y}}_{(ci)}$. It is assumed that the feature and label spaces of source and target domains are identical, but their joint probability distributions are different. Domain adaptation (DA) aims to exploit a projection to map the source and target domains data into a shared feature subspace, where their joint probability distribution divergence is minimized greatly. %\bp{Notably, DA pertains to a transductive learning paradigm where the source and target datasets are available during the model's training phase. Delete the above words?} 

\noindent \textbf{Reproducible Kernel Hilbert Space.}
A reproducible kernel Hilbert space (RKHS) is a Hilbert space ($\mathcal{H}$) of functions $f: \Omega\rightarrow \mathbf{R}$ on a domain $\Omega$, and its inner-product is $\langle \cdot,\cdot \rangle _\mathcal{H}$. The evaluation functional $f(\textbf{\textit{x}})$ can be reproduced by a kernel function $k(\textbf{\textit{x}},\textbf{\textit{x}}')$, \textit{i.e.}, $\langle f(\cdot),k(\textbf{\textit{x}},\cdot) \rangle _\mathcal{H}=f(\textbf{\textit{x}})$, and the RKHS takes its name from this so-called reproducible kernel function. Alternatively, $k(\textbf{\textit{x}},\cdot)$ can be viewed as an (infinite-dimensional) implicit map $\phi(\textbf{\textit{x}})$ where $k(\textbf{\textit{x}},\textbf{\textit{x}}')=\langle \phi(\textbf{\textit{x}}),\phi(\textbf{\textit{x}}') \rangle_\mathcal{H}$.

\subsection{Maximum Mean Discrepancy}

\par The maximum mean discrepancy (MMD) establishes the kernel mean embeddings of the marginal probability distributions in RKHS endowed by the kernel $k_\textbf{\textit{x}}$ (feature map $\varphi$) from both the source and target domains~\cite{TCA}, and computes their distance with the Hilbert-Schmidt norm as follows:
\begin{equation} 
%\footnotesize
\begin{array}{lr}
\mathbb{D}_\mathcal{H}[P_s(\textbf{\textit{X}}_s),P_t(\textbf{\textit{X}}_t)]=||\mathbb{E}[\varphi(\textbf{\textit{X}}_s)]-\mathbb{E}[\varphi(\textbf{\textit{X}}_t)]||_{\mathcal{H}}^2 
\\
\\
=||\frac{1}{n_s}\sum_{i=1}^{n_s}\varphi(\textbf{\textit{x}}_i)-\frac{1}{n_t}\sum_{j=1}^{n_t}\varphi(\textbf{\textit{x}}_j)||_{\mathcal{H}}^2
=tr(\textbf{\textit{K}}_{\textbf{\textit{x}}\textbf{\textit{x}}}\textbf{\textit{M}}_{m}\textbf{\textit{K}}_{\textbf{\textit{x}}\textbf{\textit{x}}}),
\end{array}
\label{eq1}
\end{equation}
\noindent where the kernel matrix $\textbf{\textit{K}}_{\textbf{\textit{x}}\textbf{\textit{x}}}=\varphi (\textbf{\textit{X}})^{\top}\varphi (\textbf{\textit{X}})\in \mathbf{R}^{n\times n}$ ($(\textbf{\textit{K}}_{\textbf{\textit{x}}\textbf{\textit{x}}})_{(ij)}=k(\textbf{\textit{x}}_i,\textbf{\textit{x}}_j)$), and the MMD matrix $\textbf{\textit{M}}_{m}$ can be computed as below:
\begin{equation}
(\textbf{\textit{M}}_{m})_{(ij)}=\left\{ 
\begin{array}{lr}
\frac{1}{n_sn_s},  (\textbf{\textit{x}}_i,\textbf{\textit{x}}_j\in{D_s}) &  \\
\frac{1}{n_tn_t},  (\textbf{\textit{x}}_i,\textbf{\textit{x}}_j\in{D_t}) &  \\
-\frac{1}{n_sn_t}, (otherwise). &  
\end{array}
\right .
\label{eq2}
\end{equation}

\subsection{Class-wise Maximum Mean Discrepancy}

\par The class-wise maximum mean discrepancy (CMMD) resorts to the sufficient statistics of class conditional distribution (\textit{i.e.}, $P(\textbf{\textit{X}}|\textbf{\textit{Y}})$) to approximate the conditional probability distribution (\textit{i.e.}, $P(\textbf{\textit{Y}}|\textbf{\textit{X}})$)~\cite{JDA}, which constructs the MMD for each specific class as follows:
\begin{equation} 
\footnotesize
\begin{array}{lr}
\mathbb{D}_\mathcal{H}[P_s(\textbf{\textit{X}}_s|\textbf{\textit{Y}}_s),P_t(\textbf{\textit{X}}_t|\textbf{\textit{Y}}_t)]
\\
\\
=\sum_{c=1}^{C}||\mathbb{E}[\varphi(\textbf{\textit{X}}_s^{(c)})]-\mathbb{E}[\varphi(\textbf{\textit{X}}_t^{(c)})]||_{\mathcal{H}}^2
\\
\\
=\sum_{c=1}^{C}||\frac{1}{n_s^{(c)}}\sum_{i=1}^{n_s^{(c)}}\varphi(\textbf{\textit{x}}_i^{(c)})-\frac{1}{n_t^{(c)}}\sum_{j=1}^{n_t^{(c)}}\varphi(\textbf{\textit{x}}_j^{(c)})||_{\mathcal{H}}^2
\\
\\
=\sum_{c=1}^{C}tr(\textbf{\textit{K}}_{\textbf{\textit{x}}\textbf{\textit{x}}}\textbf{\textit{M}}_{c}^{(c)}\textbf{\textit{K}}_{\textbf{\textit{x}}\textbf{\textit{x}}}),
\end{array}
\label{eq3}
\end{equation}

\noindent where the MMD matrix $\textbf{\textit{M}}_{c}^{(c)}$ can be computed as below:
\begin{equation}
(\textbf{\textit{M}}_{c}^{(c)})_{(ij)}
=\left\{ 
\begin{array}{lr}
\frac{1}{n_s^{(c)}n_s^{(c)}},  \textbf{\textit{x}}_i\in{D_s^{(c)}},\textbf{\textit{x}}_j\in{D_s^{(c)}} &  \\
\frac{1}{n_t^{(c)}n_t^{(c)}},  \textbf{\textit{x}}_i\in{D_t^{(c)}},\textbf{\textit{x}}_j\in{D_t^{(c)}} 
&  \\
-\frac{1}{n_s^{(c)}n_t^{(c)}},
\textbf{\textit{x}}_i\in{D_s^{(c)}},\textbf{\textit{x}}_j\in{D_t^{(c)}} &
\\
-\frac{1}{n_t^{(c)}n_s^{(c)}},
\textbf{\textit{x}}_i\in{D_t^{(c)}},\textbf{\textit{x}}_j\in{D_s^{(c)}}
&  \\
0, (otherwise). &  
\end{array}
\right .
\label{eq4}
\end{equation}

\subsection{Weighted Class-wise Maximum Mean Discrepancy}

\par To deal with imbalanced dataset, the weighted class-wise maximum mean discrepancy (WCMMD) introduces the class prior probability $P(\textbf{\textit{Y}})$ into the CMMD~\cite{BDA}, which pays more attention on the large-size categories and is formulated as the following:
\begin{equation} 
\begin{array}{lr}
\sum_{c=1}^{C}||\frac{P_s(\textbf{\textit{y}}_s:c)}{n_s^{(c)}}\sum_{i=1}^{n_s^{(c)}}\varphi(\textbf{\textit{x}}_i^{(c)})-\frac{P_t(\textbf{\textit{y}}_t:c)}{n_t^{(c)}}\sum_{j=1}^{n_t^{(c)}}\varphi(\textbf{\textit{x}}_j^{(c)})||_{\mathcal{H}}^2
\\
\\
=\sum_{c=1}^{C}tr(\textbf{\textit{K}}_{\textbf{\textit{x}}\textbf{\textit{x}}}\textbf{\textit{M}}_{wc}^{(c)}\textbf{\textit{K}}_{\textbf{\textit{x}}\textbf{\textit{x}}}),
\end{array}
\label{eq5}
\end{equation}

\noindent where $\textbf{\textit{M}}_{wc}^{(c)}$ can be computed with the following equation:
\begin{equation}
(\textbf{\textit{M}}_{wc}^{(c)})_{(ij)}
=\left\{ 
\begin{array}{lr}
\frac{1}{n_sn_s},  \textbf{\textit{x}}_i\in{D_s^{(c)}},\textbf{\textit{x}}_j\in{D_s^{(c)}} &  \\
\frac{1}{n_tn_t},  \textbf{\textit{x}}_i\in{D_t^{(c)}},\textbf{\textit{x}}_j\in{D_t^{(c)}} 
&  \\
-\frac{1}{n_sn_t}, 
\textbf{\textit{x}}_i\in{D_s^{(c)}},\textbf{\textit{x}}_j\in{D_t^{(c)}} &
\\
-\frac{1}{n_tn_s},
\textbf{\textit{x}}_i\in{D_t^{(c)}},\textbf{\textit{x}}_j\in{D_s^{(c)}}
&  \\
0, (otherwise). &  
\end{array}
\right .
\label{eq6}
\end{equation}

\section{Unified Joint Maximum Mean Discrepancy}
%Existing shallow DA approaches usually jointly utilize the MMD and CMDD to approximate the joint probability distribution, thus there is a need to directly establish the joint probability distribution distance to realize knowledge transfer as effective as possible. 
\par Domain adaptation aims to narrow down the joint probability distribution difference between the source and target domains, but its empirical estimation derived from the maximum mean discrepancy (joint maximum mean discrepancy, JMMD) will involve complex tensor-product operator that is hard to manipulate~\cite{JAN}. This paper unveils a unified format of JMMD that is easy to optimize. We first give the formulation of joint probability distribution in the following paragraphs.
\\ 
\noindent \textbf{Joint Probability Distribution.} With any given domain $D$ sampled from joint probability distribution $P(\textbf{\textit{X}},\textbf{\textit{Y}})$~\cite{1973}, the kernel embedding represents a joint probability distribution by an element in RKHS endowed by the kernel $k_\textbf{\textit{x}}$ (feature map $\varphi$) and $k_\textbf{\textit{y}}$ (label map $\phi$), \textit{i.e.}, uncentered co-variance operator $\mathcal{C}_{\textbf{\textit{X}}\textbf{\textit{Y}}}$. It can be formulated as below:
\begin{equation} 
\mathcal{C}_{\textbf{\textit{X}}\textbf{\textit{Y}}}:=\mathbb{E}_{\textbf{\textit{X}}\textbf{\textit{Y}}}[\phi(\textbf{\textit{x}})\otimes\varphi(\textbf{\textit{y}})],
\label{eq7}
\end{equation}

\noindent which essentially describes the feature-label correlation. Following~\cite{JAN}, we empirically estimate the embedding of joint probability distribution using finite samples, and the empirical kernel embeddings of source and target domains are as follows:
\begin{equation} 
\begin{array}{lr}
\mathcal{C}_{\textbf{\textit{X}}_s\textbf{\textit{Y}}_s}=\frac{1}{n_s}\sum_{i=1}^{n_s}[\phi(\textbf{\textit{x}}_i)\otimes\varphi(\textbf{\textit{y}}_i)],
\\
\\ \mathcal{C}_{\textbf{\textit{X}}_t\textbf{\textit{Y}}_t}=\frac{1}{n_t}\sum_{j=1}^{n_t}[\phi(\textbf{\textit{x}}_j)\otimes\varphi(\textbf{\textit{y}}_j)].
\end{array}
\label{eq8}
\end{equation}
% 在这里另起一段补充说明这两个协方差算子的物理意义.其实它们是分别对源域和目标域上特征和标签之间的关系进行建模.所以缩小联合概率分布差异实际上是使得源域和目标域上的这种关系保持一致.
\par Our goal is to reduce the feature-label correlation divergence between the two domains to draw their distributions closer. By virtue of MMD, we utilize the kernel mean embeddings of empirical joint probability distribution in RKHS of the two joint probability distributions $P_s(\textbf{\textit{X}}_s,\textbf{\textit{Y}}_s)$ and $P_t(\textbf{\textit{X}}_t,\textbf{\textit{Y}}_t)$, then compute their distance with the Hilbert-Schmidt norm. Thus, the resulting unified joint maximum mean discrepancy (JMMD) is defined as the following:
\begin{equation} 
\footnotesize
\begin{array}{lr}
\mathbb{D}_\mathcal{H}(P_s(\textbf{\textit{X}}_s,\textbf{\textit{Y}}_s),P_t(\textbf{\textit{X}}_t,\textbf{\textit{Y}}_t))
\\
\\
=||\mathbb{E}[\varphi(\textbf{\textit{X}}_s)\otimes\phi(\textbf{\textit{Y}}_s)]-\mathbb{E}[\varphi(\textbf{\textit{X}}_t)\otimes\phi(\textbf{\textit{Y}}_t)]||_{\mathcal{H}}^2 
\\
\\
=||\frac{1}{n_s}\sum_{i=1}^{n_s}[\varphi(\textbf{\textit{x}}_i)\otimes\phi(\textbf{\textit{y}}_i)]-\frac{1}{n_t}\sum_{j=1}^{n_t}[\varphi(\textbf{\textit{x}}_j)\otimes\phi(\textbf{\textit{y}}_j)]||_{\mathcal{H}}^2
\\
\\
=tr(\textbf{\textit{K}}_{\textbf{\textit{x}}\textbf{\textit{x}}}\odot \textbf{\textit{K}}_{\textbf{\textit{y}}\textbf{\textit{y}}}\textbf{\textit{M}}_{j}),
\end{array}
\label{eq9}
\end{equation}
\noindent where $\textbf{\textit{K}}_{\textbf{\textit{y}}\textbf{\textit{y}}}=\phi (\textbf{\textit{Y}})^{\top}\phi (\textbf{\textit{Y}})\in \mathbf{R}^{n\times n}$($(\textbf{\textit{K}}_{\textbf{\textit{y}}\textbf{\textit{y}}})_{(ij)}=k_\textbf{\textit{y}}(\textbf{\textit{y}}_i,\textbf{\textit{y}}_j)$), and $\textbf{\textit{M}}_{j}=\textbf{\textit{M}}_{m}$. Remarkably, in the view of Mercer kernel theorem, the nonlinear functions $\varphi$ and $\phi$ do not need to be explicit, and the complicated tensor-product disappears, more details could be found in Appendix $\textbf{\textit{A.1}}$.

\par To incorporate the unified JMMD into subspace-learning methods and leverage domain-invariant features, we formulate the unified JMMD in the projected RKHS through the following Equation:
\begin{equation} 
\begin{array}{lr}
\mathbb{D}_\mathcal{H}=||\mathbb{E}[\textbf{\textit{W}}^{\top}\varphi(\textbf{\textit{X}}_s)\otimes\phi(\textbf{\textit{Y}}_s)]-\mathbb{E}[\textbf{\textit{W}}^{\top}\varphi(\textbf{\textit{X}}_t)\otimes\phi(\textbf{\textit{Y}}_t)]||_{\mathcal{H}}^2
\\
\\
=||\frac{1}{n_s}\sum_{i=1}^{n_s}[\textbf{\textit{W}}^{\top}\varphi(\textbf{\textit{x}}_i)\otimes\phi(\textbf{\textit{y}}_i)]
\\
\\
-\frac{1}{n_t}\sum_{j=1}^{n_t}[\textbf{\textit{W}}^{\top}\varphi(\textbf{\textit{x}}_j)\otimes\phi(\textbf{\textit{y}}_j)]||_{\mathcal{H}}^2.
\end{array}
\label{eq10}
\end{equation}

\par However, the dimension of projection $\textbf{\textit{W}}\in \mathbf{R}^{\infty\times d}$ is infinite and the complicated tensor-product operator is involved in Eq.~\eqref{eq10}, so that its derivative is hard to obtain. To overcome this issue, we begin by introducing the Representer theorem.

\noindent \textbf{Theorem 1 (Representer Theorem.)} It says that any function can be decomposed into finite values of a kernel function with corresponding coefficients ~\cite{Representer,ARTL}.

\begin{equation} 
\begin{array}{lr}
\textbf{\textit{W}}^{\top}\varphi(\textbf{\textit{x}})=\sum_{i=1}^{n_{st}}\textbf{\textit{b}}_ik(\textbf{\textit{x}},\textbf{\textit{x}}_i)=\sum_{i=1}^{n_{st}}\textbf{\textit{b}}_i\langle \varphi(\textbf{\textit{x}}),\varphi(\textbf{\textit{x}}_i) \rangle 
\\
\\
=\sum_{i=1}^{n_{st}}\textbf{\textit{b}}_i\varphi(\textbf{\textit{x}}_i)^{\top}\varphi(\textbf{\textit{x}}),
\end{array}
\label{eq11}
\end{equation}
\noindent where $\textbf{\textit{b}}_i\in\mathbf{R}^{d\times1}$, %and
thus we definite a new projection matrix $\textbf{\textit{B}}=[\textbf{\textit{b}}_1^{\top};...;\textbf{\textit{b}}_{n_{st}}^{\top}]\in\mathbf{R}^{n_{st}\times d}$. According to Eq.~\eqref{eq11}, then Eq.~\eqref{eq10} could be rewritten to:
\begin{equation} 
\begin{array}{lr}
\mathbb{D}_\mathcal{H}=||\frac{1}{n_s}\sum_{i=1}^{n_s}[\{\sum_{l=1}^{n_{st}}\textbf{\textit{b}}_l\varphi(\textbf{\textit{x}}_l)^{\top}\}\varphi(\textbf{\textit{x}}_i)\otimes\varphi(\textbf{\textit{y}}_i)]
\\
\\
-\frac{1}{n_t}\sum_{j=1}^{n_t}[\{\sum_{l=1}^{n_{st}}\textbf{\textit{b}}_l\varphi(\textbf{\textit{x}}_l)^{\top}\}\varphi(\textbf{\textit{x}}_j)\otimes\varphi(\textbf{\textit{y}}_j)]||_{\mathcal{H}}^2
\\
\\
=tr(\{\textbf{\textit{K}}_{\textbf{\textit{x}}\textbf{\textit{x}}}\textbf{\textit{B}}\textbf{\textit{B}}^{\top}\textbf{\textit{K}}_{\textbf{\textit{x}}\textbf{\textit{x}}}\}\odot \textbf{\textit{K}}_{\textbf{\textit{y}}\textbf{\textit{y}}}\textbf{\textit{M}}_{j}).
\end{array}
\label{eq12}
\end{equation}

\noindent Remarkably, the matrix $\textbf{\textit{W}}$ does not need to optimize because we resort to optimize $\textbf{\textit{B}}$ instead, and the tensor-product operator also disappears, more details could be found in Appendix $\textbf{\textit{A.2}}$.

\section{New Discoveries from the Unified JMMD} 
\subsection{A Novel MMD Matrix}
\par The Hilbert Schmidt independence (HSI)~\cite{HSI} also utilizes Eq.~\eqref{eq7} to establish the feature-label correlation. Different from JMMD, HSI aims to maximize the correlation for a given domain to improve its feature discriminability, but JMMD minimizes the correlation bias between the two domains to reduce their distribution distance. An obvious problem is that the domain-specific correlation (discriminability) may be reduced unexpectedly during the process of JMMD minimization (transferability), and this problem will be demonstrated in Section of~\ref{Abation Study}. 
Our previous unified JMMD analysis provides a theoretical basis about this contradiction.
%This contradiction provides a theoretical viewpoint to illustrate that the relationship between the feature transferability and its discriminability is as one falls, another rises. 
Therefore, we should simultaneously minimize the JMMD and maximize the domain-specific HSI. Specially, the domain-specific HSI metrics for the source and target domains can be formulated as follows,
%\par As can be observed that the unified JMMD (\ie, Eq. (\ref{eq9})) aims to minimize the feature-label correlation divergence across the source and target domains, while the domain-specific correlation may be reduced unexpectedly and the feature discriminability is degraded which benefits to classification (it is verified in section of~\ref{Abation Study}). To this end, we should simultaneously minimize the correlation divergence and maximize the domain-specific correlation. According to the Hilbert Schmidt independence (HSI)~\cite{HSI}, the domain-specific feature-label covariance for the source and target domains can be formulated as follows:

\begin{equation} 
\begin{array}{lr}
||\mathcal{C}_{\textbf{\textit{X}}_s\textbf{\textit{Y}}_s}||_{\mathcal{H}}^2+||\mathcal{C}_{\textbf{\textit{X}}_t\textbf{\textit{Y}}_t}||_{\mathcal{H}}^2=||\frac{1}{n_s}\sum_{i=1}^{n_s}[\varphi(\textbf{\textit{x}}_i)\otimes\varphi(\textbf{\textit{y}}_i)]||_{\mathcal{H}}^2+
\\
\\
||\frac{1}{n_t}\sum_{i=1}^{n_t}[\varphi(\textbf{\textit{x}}_i)\otimes\varphi(\textbf{\textit{y}}_i)]||_{\mathcal{H}}^2=tr((\textbf{\textit{K}}_{\textbf{\textit{x}}\textbf{\textit{x}}})_s\odot(\textbf{\textit{K}}_{\textbf{\textit{y}}\textbf{\textit{y}}})_s\textbf{\textit{M}}_{hs})
\\
\\
+tr((\textbf{\textit{K}}_{\textbf{\textit{x}}\textbf{\textit{x}}})_t\odot(\textbf{\textit{K}}_{\textbf{\textit{y}}\textbf{\textit{y}}})_t\textbf{\textit{M}}_{ht})=tr(\textbf{\textit{K}}_{\textbf{\textit{x}}\textbf{\textit{x}}}\odot\textbf{\textit{K}}_{\textbf{\textit{y}}\textbf{\textit{y}}}\textbf{\textit{M}}_h).
\end{array}
\label{eq13}
\end{equation}

\par Similarly, we can utilize the Representer theorem to obtain its variant in the projected RKHS, and it can be formulated as below:
\begin{equation} 
\begin{array}{lr}
tr(\{\textbf{\textit{K}}_{\textbf{\textit{x}}\textbf{\textit{x}}}\textbf{\textit{B}}\textbf{\textit{B}}^{\top}\textbf{\textit{K}}_{\textbf{\textit{x}}\textbf{\textit{x}}}\}\odot \textbf{\textit{K}}_{\textbf{\textit{y}}\textbf{\textit{y}}}\textbf{\textit{M}}_{h}),
\end{array}
\label{eq14}
\end{equation}

\noindent where $\textbf{\textit{M}}_{h}$ can be computed as the following equation:
\begin{equation}
(\textbf{\textit{M}}_{h})_{(ij)}=\left\{ 
\begin{array}{lr}
\frac{1}{n_sn_s},  (\textbf{\textit{x}}_i,\textbf{\textit{x}}_j\in{D_s}) &  \\
\frac{1}{n_tn_t},  (\textbf{\textit{x}}_i,\textbf{\textit{x}}_j\in{D_t}) &  \\
0, (otherwise). &  
\end{array}
\right .
\label{eq15}
\end{equation}

Different from ~\cite{HSI,ACA,ACA2,CSRTL,MIDA}, we utilize the HSI with its uncentered variant for consistency with the JMMD, and combine the JMMD and HSI in a neat format with a novel MMD matrix, \textit{i.e.}, $(\textbf{\textit{M}}_{j}-\delta\textbf{\textit{M}}_{h})$. Thus, the first discovery from the unified JMMD to improve its feature discriminability 
%proposed robust JMMD with a novel MMD matrix (\ie, $\textbf{\textit{M}}_{j}-\delta\textbf{\textit{M}}_{h}$)) 
is finalized as,
\begin{equation} 
\begin{array}{lr}
tr(\textbf{\textit{K}}_{\textbf{\textit{x}}\textbf{\textit{x}}}\odot \textbf{\textit{K}}_{\textbf{\textit{y}}\textbf{\textit{y}}}(\textbf{\textit{M}}_{j}-\delta\textbf{\textit{M}}_{h})),
\end{array}
\label{eq16}
\end{equation}

\begin{equation} 
\begin{array}{lr}
tr(\{\textbf{\textit{K}}_{\textbf{\textit{x}}\textbf{\textit{x}}}\textbf{\textit{B}}\textbf{\textit{B}}^{\top}\textbf{\textit{K}}_{\textbf{\textit{x}}\textbf{\textit{x}}}\}\odot \textbf{\textit{K}}_{\textbf{\textit{y}}\textbf{\textit{y}}}(\textbf{\textit{M}}_{j}-\delta\textbf{\textit{M}}_{h})),
\end{array}
\label{eq17}
\end{equation}
\noindent where $\delta$ is a trade-off parameter to balance their importance.

\subsection{A Novel Label Kernel}
\noindent \textbf{Theorem 2} The MMD, CMMD and WCMMD are the special cases of the proposed unified JMMD with different label kernels, \textit{i.e.}, $\textbf{\textit{K}}_{\textbf{\textit{y}}\textbf{\textit{y}}}^1=\textbf{\textit{1}}_{n_{st}\times n_{st}}$, $\textbf{\textit{K}}_{\textbf{\textit{y}}\textbf{\textit{y}}}^2$ and $\textbf{\textit{K}}_{\textbf{\textit{y}}\textbf{\textit{y}}}^3$. The $\textbf{\textit{K}}_{\textbf{\textit{y}}\textbf{\textit{y}}}^2$ and $\textbf{\textit{K}}_{\textbf{\textit{y}}\textbf{\textit{y}}}^3$ are defined as follows:
\begin{equation}
(\textbf{\textit{K}}_{\textbf{\textit{y}}\textbf{\textit{y}}}^2)_{(ij)}
=\left\{ 
\begin{array}{lr}
\frac{n_sn_s}{n_s^{(c)}n_s^{(c)}},  \textbf{\textit{x}}_i\in{D_s^{(c)}},\textbf{\textit{x}}_j\in{D_s^{(c)}} &  \\
\frac{n_tn_t}{n_t^{(c)}n_t^{(c)}}, \textbf{\textit{x}}_i\in{D_t^{(c)}},\textbf{\textit{x}}_j\in{D_t^{(c)}} 
&  \\
\frac{n_sn_t}{n_s^{(c)}n_t^{(c)}},
\textbf{\textit{x}}_i\in{D_s^{(c)}},\textbf{\textit{x}}_j\in{D_t^{(c)}} &
\\
\frac{n_tn_s}{n_t^{(c)}n_s^{(c)}},
\textbf{\textit{x}}_i\in{D_t^{(c)}},\textbf{\textit{x}}_j\in{D_s^{(c)}}
&  \\
0, (otherwise). &  
\end{array}
\right .
\label{eq18}
\end{equation}

\begin{equation}
(\textbf{\textit{K}}_{\textbf{\textit{y}}\textbf{\textit{y}}}^3)_{(ij)}
=\left\{ 
\begin{array}{lr}
1,  \textbf{\textit{x}}_i\in{D_s^{(c)}},\textbf{\textit{x}}_j\in{D_s^{(c)}} &  \\
1, \textbf{\textit{x}}_i\in{D_t^{(c)}},\textbf{\textit{x}}_j\in{D_t^{(c)}} 
&  \\
1, 
\textbf{\textit{x}}_i\in{D_s^{(c)}},\textbf{\textit{x}}_j\in{D_t^{(c)}} &
\\
1,
\textbf{\textit{x}}_i\in{D_t^{(c)}},\textbf{\textit{x}}_j\in{D_s^{(c)}}
&  \\
0, (otherwise). &  
\end{array}
\right .
\label{eq19}
\end{equation}

\par The proof of Theorem 2 could be found in Appendix $\textbf{\textit{A.4}}$, and we also prove that the label kernels formulated here are reproducible in Appendix $\textbf{\textit{A.3}}$.
% 交待怎么引出这两个发现的
%\bp{Add more explanations about what kind of benefits or advantages can this unified frame bring.}
% 这里的不同核矩阵对应已有不同的分布距离,说明了所提统一模型的灵活性和可扩展性,方便我们设计更多有意思的标签核,从而来提高这个统一的分布距离针对不同实际问题的有效性和鲁棒性,如我们接下来将会引出一种新的标签核矩阵来应对标签分布漂移的问题.此外,这个统一的模型可以方便我们将JMMD和HSI联系起来,阐述二者此消彼长的关系对整个知识迁移产生的影响,如接下来我们将提出一个新的MMD矩阵来同时考虑JMMD和HSI.
%\section{Robust joint maximum mean discrepancy} 
%% 这里需要对HSI的动机描述的有深度一点 给Novel MMD matrix 取名字
% 取个好听的名字
% \subsection{A Novel Label Kernel}
\par Notably, Theorem 2 yields a better guidance to refine the JMMD by devising more delicate label kernels for different problems in domain adaptation. Specially, from Eq.~\eqref{eq5}, the class conditional probability distributions across domains will be mismatched when $P(\textbf{\textit{Y}}_s)\neq P(\textbf{\textit{Y}}_t)$, and this problem will be demonstrated in Section of~\ref{Abation Study}. Inspired by~\cite{WMMD}, we re-weight a source data instance $\textbf{\textit{x}}_i$ that pertains to $c$-th category with $P(\textbf{\textit{y}}_t:c)/P(\textbf{\textit{y}}_s:c)$, so that the label distribution shift is alleviated. Accordingly, a novel label kernel can be proposed as below:
\begin{equation}
(\textbf{\textit{K}}_{\textbf{\textit{y}}\textbf{\textit{y}}}^4)_{(ij)}
=\left\{ 
\begin{array}{lr}
\frac{n_t^{(c)}n_t^{(c)}n_sn_s}{n_tn_tn_s^{(c)}n_s^{(c)}},  \textbf{\textit{x}}_i\in{D_s^{(c)}},\textbf{\textit{x}}_j\in{D_s^{(c)}} &  \\
1, \textbf{\textit{x}}_i\in{D_t^{(c)}},\textbf{\textit{x}}_j\in{D_t^{(c)}} 
&  \\
\frac{n_t^{(c)}n_s}{n_tn_s^{(c)}},
\textbf{\textit{x}}_i\in{D_s^{(c)}},\textbf{\textit{x}}_j\in{D_t^{(c)}} &
\\
\frac{n_t^{(c)}n_s}{n_tn_s^{(c)}},
\textbf{\textit{x}}_i\in{D_t^{(c)}},\textbf{\textit{x}}_j\in{D_s^{(c)}}
&  \\
0, (otherwise). &  
\end{array}
\right .
\label{eq20}
\end{equation}
% 在这个公式下多说几句话.
\noindent  We also prove that $\textbf{\textit{K}}_{\textbf{\textit{y}}\textbf{\textit{y}}}^4$ is reproducible in Appendix $\textbf{\textit{A.3}}$.
\subsection{Applications}

\par To validate the effectiveness of the proposed two discoveries from the unveiled unified JMMD, we leverage it into the traditional subspace-learning method to embody the robustness and effectiveness in terms of extracting domain-invariant features. For simplicity, we introduce them into the principle component analysis (PCA) framework as:
% 这里一是为了简单,二是为了用过复杂的特征提取模型会有其它因素的干扰,不能体现所提理论结果的有效性.
\begin{equation} 
\begin{array}{lr}
\min\limits_{\textbf{\textit{B}}}tr(\{\textbf{\textit{K}}_{\textbf{\textit{x}}\textbf{\textit{x}}}\textbf{\textit{B}}\textbf{\textit{B}}^{\top}\textbf{\textit{K}}_{\textbf{\textit{x}}\textbf{\textit{x}}}\}\odot \textbf{\textit{K}}_{\textbf{\textit{y}}\textbf{\textit{y}}}(\textbf{\textit{M}}_{j}-\delta\textbf{\textit{M}}_{h}))+
\\
\\
\lambda||\textbf{\textit{B}}||_F^2
\quad s.t. \quad  \textbf{\textit{B}}^{\top}\textbf{\textit{K}}_{\textbf{\textit{x}}\textbf{\textit{x}}}\textbf{\textit{H}}\textbf{\textit{K}}_{\textbf{\textit{x}}\textbf{\textit{x}}}\textbf{\textit{B}}=\textbf{\textit{I}}_d,
\end{array}
\label{eq21}
\end{equation}
\noindent where $\lambda$ is a trade-off parameter and we can utilize different label kernels including the proposed $\textbf{\textit{K}}_{\textbf{\textit{y}}\textbf{\textit{y}}}^4$, thus we can learn more effective domain-invariant features with the above two discoveries.

\par Similar to previous works~\cite{JDA,TJM}, we can obtain a generalized eigendecomposition problem as:
\begin{equation} 
\begin{array}{lr}
(\textbf{\textit{K}}_{\textbf{\textit{x}}\textbf{\textit{x}}}\{(\textbf{\textit{M}}_{j}-\delta\textbf{\textit{M}}_{h})\odot\textbf{\textit{K}}_{\textbf{\textit{y}}\textbf{\textit{y}}}\}\textbf{\textit{K}}_{\textbf{\textit{x}}\textbf{\textit{x}}}
+\lambda\textbf{\textit{I}}_m)\textbf{\textit{B}}=\textbf{\textit{K}}_{\textbf{\textit{x}}\textbf{\textit{x}}}\textbf{\textit{H}}\textbf{\textit{K}}_{\textbf{\textit{x}}\textbf{\textit{x}}}\textbf{\textit{B}}\Theta,
\end{array}
\label{eq22}
\end{equation}
\noindent where $\Theta\in {\mathbf{R}^{d\times d}}$ is a diagonal matrix with Lagrange Multipliers, and Eq.~\eqref{eq22} can be solved by calculating the eigenvectors corresponding to the $d$-smallest eigenvalues.
% 这说明我们所提的理论用到子空间学习方法中是非常方便优化的.
%-------------------------------------------------------------------------
\section{Experiments}

\subsection{Datasets and Experimental Settings}
\par To demonstrate the validity of our theoretical results, we conducted experiments on the following 3 benchmark datasets in cross-domain object recognition. \textbf{Office10-Caltech10}\footnote{https://github.com/jindongwang/transferlearning/tree/master/data} consists of 4 domains (\textit{i.e.}, Amazon, Dslr, Webcam, Caltech) and 10 categories are shared. \textbf{Office31}\textsuperscript{\ref{l1}} contains 3 domains (\textit{i.e.}, Amazon, Dslr, Webcam) and 31 categories are shared. \textbf{Office-Home}\footnote{https://github.com/hellowangqian/domainadaptation-capls\label{l1}} has 4 domains and 65 common categories.

\begin{table*}[t]
	\caption{Standard domain adaption (left part) and label distribution shift (right part) based classification results on Office10-Caltech10.}
	\label{tab1}
	\begin{center}
		\begin{tabu}{|c|c|c|c|c|c|c|c|c|[2pt]c|c|c|c|}
			\hline
			S & T & KNN & M & M$^{*}$ & C & C$^{*}$ & WC & WC$^{*}$ & KNN & WC & WWC & WWC$^{*}$ \\ 
			\hline\hline
			\multirow{3}*{C} & A & 36.0 & 46.6 & 47.1 & 46.1 & 50.9 & 44.9 & 50.4 & 32.3\small{$\pm$6.0} & 37.4\small{$\pm$6.2} & 39.0\small{$\pm$7.2} & 43.4\small{$\pm$8.0} \\
			\cline{2-13}
			& W & 29.2 & 37.3 & 37.3 & 42.4 & 43.4 & 41.0 & 44.4 & 29.1\small{$\pm$7.2} & 32.9\small{$\pm$5.8} & 34.5\small{$\pm$6.5} & 37.4\small{$\pm$6.7} \\
			\cline{2-13}
			& D & 38.2 & 46.5 & 47.1 & 43.3 & 49.0 & 42.7 & 50.3 & 37.6\small{$\pm$4.2} & 37.8\small{$\pm$5.1} & 39.5\small{$\pm$4.7} & 44.4\small{$\pm$4.3} \\
			\hline
			\multirow{3}*{A} & C & 34.2 & 38.9 & 39.9 & 39.1 & 42.4 & 38.6 & 42.3 & 29.8\small{$\pm$5.4} & 32.0\small{$\pm$6.2} & 32.4\small{$\pm$7.2} & 33.9\small{$\pm$7.2} \\
			\cline{2-13}
			& W & 31.2 & 39.7 & 39.7 & 39.0 & 45.8 & 45.8 & 46.4 & 31.8\small{$\pm$3.6} & 40.9\small{$\pm$5.2} & 41.7\small{$\pm$5.8} & 43.3\small{$\pm$5.7} \\
			\cline{2-13}
			& D & 35.7 & 38.2 & 39.5 & 41.4 & 46.5 & 38.9 & 45.9 & 27.5\small{$\pm$7.9} & 36.7\small{$\pm$7.2} & 38.0\small{$\pm$7.3} & 39.6\small{$\pm$7.7} \\
			\hline
			\multirow{3}*{W} & C & 28.8 & 31.3 & 31.3 & 32.9 & 35.2 & 33.1 & 35.4 & 24.4\small{$\pm$3.8} & 26.7\small{$\pm$3.3} & 27.1\small{$\pm$3.2} & 28.7\small{$\pm$4.1} \\
			\cline{2-13}
			& A & 31.6 & 29.5 & 30.3 & 40.3 & 41.4 & 40.8 & 41.6 & 27.9\small{$\pm$4.3} & 33.6\small{$\pm$6.9} & 36.2\small{$\pm$5.7} & 37.1\small{$\pm$5.8} \\
			\cline{2-13}
			& D & 84.7 & 90.4 & 90.4 & 90.4 & 94.3 & 89.8 & 92.4 & 70.1\small{$\pm$8.5} & 68.6\small{$\pm$5.8} & 73.2\small{$\pm$6.1} & 76.0\small{$\pm$8.0} \\
			\hline
			\multirow{3}*{D} & C & 29.6 & 33.1 & 33.1 & 30.0 & 32.4 & 31.1 & 33.9 & 28.0\small{$\pm$5.8} & 32.1\small{$\pm$6.3} & 32.0\small{$\pm$6.3} & 33.5\small{$\pm$6.3} \\
			\cline{2-13}
			& A & 28.3 & 33.6 & 33.8 & 35.3 & 39.0 & 37.0 & 40.0 & 25.9\small{$\pm$3.8} & 30.1\small{$\pm$5.5} & 30.2\small{$\pm$5.8} & 31.6\small{$\pm$5.6} \\
			\cline{2-13}
			& W & 83.7 & 88.5 & 88.5 & 91.5 & 92.9 & 90.2 & 91.5 & 69.2\small{$\pm$7.9} & 72.4\small{$\pm$8.2} & 75.4\small{$\pm$8.1} & 76.7\small{$\pm$8.0} \\
			\hline
			& Avg. & 40.9 & 46.1 & \textbf{46.5} & 47.6 & \textbf{51.1} & 47.8 & \textbf{51.2} & 36.1\small{$\pm$5.7} & 40.1\small{$\pm$6.0} & \textbf{41.6}\small{$\pm$6.2} & \textbf{43.8}\small{$\pm$6.5} \\
			\hline
		\end{tabu}
	\end{center}
\end{table*}

\begin{table*}[h]
	\caption{Standard domain adaption (left part) and label distribution shift (right part) based classification results on Office31.}
	\label{tab2}
	% 	\vspace{-6pt}
	\begin{center}
		\begin{tabu}{|c|c|c|c|c|c|c|c|c|[2pt]c|c|c|c|}
			\hline
			S & T & KNN & M & M$^{*}$ & C & C$^{*}$ & WC & WC$^{*}$ & KNN & WC & WWC & WWC$^{*}$ \\
			\hline\hline
			\multirow{2}*{A} & D & 78.1 & 77.5 & 77.5 & 79.9 & 82.1 & 79.7 & 83.7 & 75.9\small{$\pm$4.7} & 75.5\small{$\pm$4.3} & 78.0\small{$\pm$3.5} & 80.4\small{$\pm$2.9} \\
			\cline{2-13}
			& W & 78.0 & 78.4 & 78.4 & 82.3 & 84.5 & 82.8 & 84.2 & 72.0\small{$\pm$4.6} & 71.1\small{$\pm$3.9} & 74.0\small{$\pm$4.7} & 74.9\small{$\pm$4.5} \\
			\hline
			\multirow{2}*{D} & A & 66.8 & 67.4 & 67.4 & 68.7 & 69.7 & 68.7 & 69.7 & 65.4\small{$\pm$2.8} & 66.4\small{$\pm$3.1} & 67.3\small{$\pm$2.3} & 68.1\small{$\pm$2.3} \\
			\cline{2-13}
			& W & 98.5 & 98.7 & 98.7 & 98.9 & 98.9 & 98.5 & 98.9 & 92.3\small{$\pm$3.4} & 91.6\small{$\pm$3.2} & 93.0\small{$\pm$3.2} & 93.2\small{$\pm$3.1} \\
			\hline
			\multirow{2}*{W} & A & 63.9 & 64.6 & 64.6 & 67.1 & 69.1 & 67.1 & 67.2 & 61.4\small{$\pm$2.2} & 62.3\small{$\pm$2.4} & 63.8\small{$\pm$2.3} & 64.1\small{$\pm$2.3} \\
			\cline{2-13}
			& D & 99.4 & 99.4 & 99.4 & 99.4 & 100.0 & 99.2 & 99.4 & 90.1\small{$\pm$3.8} & 88.7\small{$\pm$3.6} & 90.6\small{$\pm$3.8} & 91.3\small{$\pm$4.0} \\
			\hline
			& Avg. & 80.8 & 81.0 & \textbf{81.0} & 82.7 & \textbf{84.1} & 82.7 & \textbf{83.9} & 76.2\small{$\pm$3.6} & 75.9\small{$\pm$3.4} & \textbf{77.8}\small{$\pm$3.3} & \textbf{78.7}\small{$\pm$3.2} \\
			\hline
		\end{tabu}
	\end{center}
\end{table*}

\par The proposed two strategies only involve one hyper-parameter $\delta$. After trials, we set $\delta=0.1$ on the Office10-Caltech10 and $\delta=0.5$ on the other two datasets, More specifically, we further provide empirical analysis on its sensitivity on  Office10-Caltech10 in Section \ref{sensitivity}. We adopt both the shallow and deep features as the inputs of Eq.~\eqref{eq21}. On Office10-Caltech10, the SURF features with 800 dimensions are adopted~\cite{GFK,JDA}. On the other two datasets, we utilize the deep features with 2048 dimensions extracted from the ResNet-50 model pre-trained on ImageNet~\cite{ResNet,SP}. We simulate label distribution shift by randomly dropping out 50\% samples in the ﬁrst half of classes within source domain, and 50\% samples in latter half of classes in target domain~\cite{MLD}, then random selection is repeated 10 times and average results and standard deviations are reported. More details about our implementation could be found in Appendix $\textbf{\textit{B}}$.  

\begin{figure}[t]
	\begin{center}
		\includegraphics[width=1.0\linewidth]{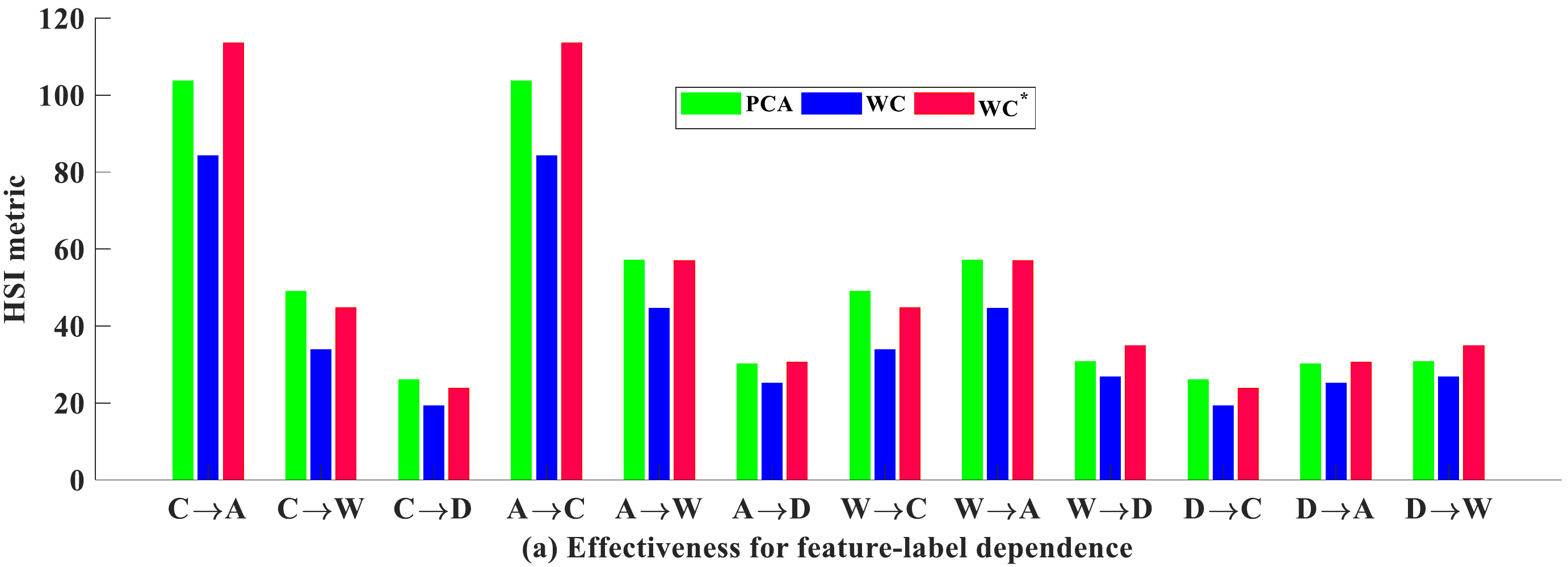}
		\includegraphics[width=1.0\linewidth]{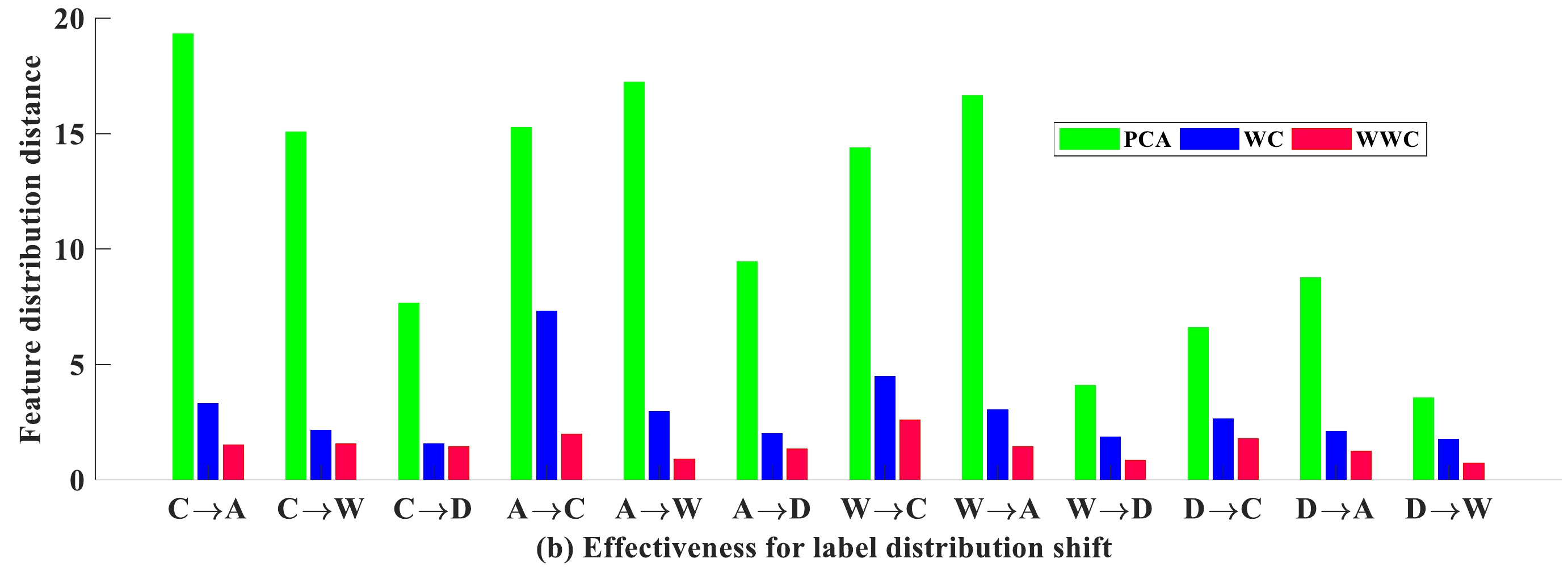}
	\end{center}
	\caption{Correlation in HSI and feature distribution distance for PCA, WC and WWC. (a) feature-label correlation for different task tests. (b) feature distribution distance for different task tests.}
	\label{fig2}
	% \vspace{-6pt}
\end{figure}

\subsection{Results}

\par We utilize the label kernels $\textbf{\textit{K}}_{yy}^1$, $\textbf{\textit{K}}_{yy}^2$, $\textbf{\textit{K}}_{yy}^3$, then the proposed JMMD is degenerated to the marginal, class conditional and weighted class conditional distribution distances (\textit{i.e.}, M, C, WC). To validate the effectiveness of feature-label correlation reinforcement strategy, we introduce the proposed MMD matrix into them (\textit{i.e.}, M$^*$, C$^*$, WC$^*$) and obtain the new feature representations by Eq. \eqref{eq22}, \textit{i.e.}, $\textbf{\textit{B}}^{\top}\textbf{\textit{K}}_{xx}$. We also replace $\textbf{\textit{K}}_{yy}^3$ in WC and WC$^*$ with the proposed label kernel $\textbf{\textit{K}}_{yy}^4$ and represent their corresponding variants with WWC, WWC$^*$. Notably, the parameters $\lambda$, $d$, $T$ are set as $0.1,20,5$ for Office10-Caltech10 and $1.0,100,5$ for the others~\cite{JDA}, and we adopt the classifier k-neareast neighbors (KNN) to show the effects of domain adaption. %for simplicity. 

\par As can be seen from the left parts of Tables~\ref{tab1},~\ref{tab2},~\ref{tab3}, the results of average accuracy of M$^*$, C$^*$, WC$^*$ are 46.5\%, 51.1\%, 51.2\% / 81.0\%, 84.1\%, 83.9\% / 58.1\%, 61.6\%, 60.4\%, respectively, which validates its capability (the proposed novel MMD matrix for feature-label dependency or feature discriminability) comparing with those corresponding results 46.1\%, 47.6\%, 47.8\% / 81.0\%, 82.7\%, 82.7\% / 58.0\%, 58.9\%, 58.1\%.
From right parts of Tables~\ref{tab1},~\ref{tab2},~\ref{tab3}, the results of average accuracy of WWC and WWC$^*$ are 41.6\%, 43.8\% / 77.8\%, 78.7\% / 52.1\%, 53.5\% which validates its robustness for label distribution shift comparing with those corresponding results of WC 40.1\% / 75.9\% / 50.0\%.   
% 说明一下不加深度方法的原因
\par To further show the effectiveness of the proposed strategies for the unified JMMD, we chose some typical classifiers and some DA methods. As such we incorporate them into other classifiers such as (\textit{i.e.}, SVM\footnote{http://www.csie.ntu.edu.tw/~cjlin/libsvm}, LP~\cite{SODA,GGSL}, NCP~\cite{SP}) and existing popular domain adaptation approaches (\textit{i.e.}, VDA~\cite{VDA}, EasyTL~\cite{easyTL}, SP~\cite{SP}). We compare WC and WC$^*$ in standard DA (left part) and WC and WWC$^*$ in label distribution shift (right part) for each method, and the results are recorded in Table~\ref{tab4} and Table~\ref{tab5}, respectively. We can notice that the proposed two strategies also works well for other classifiers since it can increase the classification accuracy a lot for all three benchmark datasets. As shown in Table ~\ref{tab5}, the proposed two strategies for JMMD can also improve these typical DA methods by showing a large margin of accuracy over all the three benchmark datasets (D$_1$: Office10-Caltech10, D$_2$: Office31, D$_3$: Office-Home, the average results for 12, 6, 12 DA tasks are reported). It should be noted that to show fairness, we did NOT compare performance of the proposed unified JMMD to some latest works that are based on deep learning methods such as ~\cite{CVPR4,CAN,CVPR5,CVPR6} and so on.

\begin{table*}[t]
	\caption{Standard domain adaption (left part) and label distribution shift (right part) based classification results on Office-Home.}
	\label{tab3}
	\begin{center}
		\begin{tabu}{|c|c|c|c|c|c|c|c|c|[2pt]c|c|c|c|}
			\hline
			S & T & KNN & M & M$^{*}$ & C & C$^{*}$ & WC & WC$^{*}$ & KNN & WC & WWC & WWC$^{*}$ \\ 
			\hline\hline
			\multirow{3}*{A} & C & 42.7 & 42.9 & 42.9 & 44.3 & 47.8 & 42.9 & 45.6 & 38.1\small{$\pm$2.8} & 38.1\small{$\pm$3.0} & 39.0\small{$\pm$2.8} & 40.5\small{$\pm$3.2} \\
			\cline{2-13}
			& P & 59.9 & 61.6 & 61.6 & 62.3 & 65.2 & 60.8 & 64.4 & 51.8\small{$\pm$2.3} & 51.7\small{$\pm$2.7} & 54.5\small{$\pm$2.6} & 55.7\small{$\pm$3.1} \\
			\cline{2-13}
			& R & 65.7 & 66.3 & 66.6 & 67.0 & 69.8 & 66.5 & 68.2 & 57.4\small{$\pm$2.4} & 57.6\small{$\pm$2.5} & 58.7\small{$\pm$2.7} & 60.2\small{$\pm$2.3} \\
			\hline
			\multirow{3}*{C} & A & 50.6 & 51.6 & 51.5 & 52.9 & 55.7 & 50.8 & 53.9 & 42.8\small{$\pm$2.9} & 41.1\small{$\pm$2.6} & 44.8\small{$\pm$2.6} & 45.9\small{$\pm$2.7} \\
			\cline{2-13}
			& P & 58.3 & 58.1 & 58.2 & 59.9 & 63.8 & 58.7 & 62.2 & 51.8\small{$\pm$2.8} & 50.9\small{$\pm$2.7} & 53.0\small{$\pm$2.9} & 54.8\small{$\pm$3.4} \\
			\cline{2-13}
			& R & 61.2 & 62.0 & 62.0 & 61.8 & 65.3 & 61.0 & 63.5 & 50.6\small{$\pm$2.3} & 50.0\small{$\pm$2.2} & 52.4\small{$\pm$2.4} & 54.1\small{$\pm$1.9} \\
			\hline
			\multirow{3}*{P} & A & 53.6 & 53.2 & 53.3 & 54.4 & 57.8 & 53.8 & 57.2 & 47.1\small{$\pm$1.9} & 46.1\small{$\pm$1.8} & 48.2\small{$\pm$2.3} & 49.6\small{$\pm$ 2.0} \\
			\cline{2-13}
			& C & 44.4 & 44.5 & 44.4 & 46.0 & 48.3 & 45.6 & 47.2 & 39.8\small{$\pm$1.9} & 40.1\small{$\pm$2.5} & 40.8\small{$\pm$2.3} & 42.0\small{$\pm$2.4} \\
			\cline{2-13}
			& R & 70.5 & 70.3 & 70.5 & 71.2 & 72.6 & 71.0 & 72.4 & 64.0\small{$\pm$2.5} & 63.8\small{$\pm$2.5} & 65.7\small{$\pm$2.2} & 66.4\small{$\pm$2.0} \\
			\hline
			\multirow{3}*{R} & A & 61.5 & 62.2 & 62.3 & 62.6 & 64.4 & 61.4 & 63.5 & 55.1\small{$\pm$2.1} & 53.3\small{$\pm$2.5} & 56.2\small{$\pm$2.5} & 58.0\small{$\pm$2.6} \\
			\cline{2-13}
			& C & 48.4 & 48.3 & 48.3 & 49.9 & 51.6 & 49.4 & 50.7 & 42.7\small{$\pm$1.6} & 41.8\small{$\pm$2.2} & 43.5\small{$\pm$1.7} & 45.0\small{$\pm$1.5} \\
			\cline{2-13}
			& P & 74.9 & 75.1 & 75.1 & 75.1 & 76.4 & 74.9 & 75.7 & 67.0\small{$\pm$1.3} & 65.2\small{$\pm$1.6} & 68.3\small{$\pm$1.3} & 69.7\small{$\pm$1.6} \\
			\hline
			& Avg. & 57.6 & 58.0 & \textbf{58.1} & 58.9 & \textbf{61.6} & 58.1 & \textbf{60.4} & 50.7\small{$\pm$2.2} & 50.0\small{$\pm$2.4} & \textbf{52.1}\small{$\pm$2.4} & \textbf{53.5}\small{$\pm$2.4} \\
			\hline
		\end{tabu}
	\end{center}
\end{table*}

\begin{table*}[h]
	\caption{Standard domain adaption (left part) and label distribution shift (right part) based classification results for typical classifiers.}
	\label{tab4}
	\begin{center}
		\begin{tabu}{|c|c|c|[2pt]c|c|[2pt]c|c|[2pt]c|c|[2pt]c|c|[2pt]c|c|}
			\hline
			{} & \multicolumn{4}{|c|[2pt]}{SVM} & \multicolumn{4}{|c|[2pt]}{LP} & \multicolumn{4}{|c|}{NCP} \\ 
			\hline
			%\hline
			{} & WC & WC$^{*}$ & WC & WWC$^{*}$ & WC & WC$^{*}$ & WC & WWC$^{*}$ & WC & WC$^{*}$ & WC & WWC$^{*}$ \\ 
			\cline{1-13}
			D$_1$ & 48.7 & 52.0 & 38.3\small{$\pm$7.8} & 42.3\small{$\pm$7.2} & 53.0 & 55.0 & 40.4\small{$\pm$7.2} & 42.6\small{$\pm$7.5} & 47.1 & 50.0 & 40.3\small{$\pm$7.4} & 43.4\small{$\pm$7.3} \\
			\hline
			D$_2$ & 82.2 & 83.0 & 64.4\small{$\pm$6.1} & 66.2\small{$\pm$6.1} & 85.7 & 87.3 & 76.3\small{$\pm$5.2} & 79.6\small{$\pm$4.8} & 83.5 & 84.8 & 75.4\small{$\pm$4.9} & 78.6\small{$\pm$5.1} \\
			\hline
			D$_3$ & 62.8 & 64.1 & 48.3\small{$\pm$3.4} & 50.2\small{$\pm$3.4} & 62.8 & 64.7 & 52.6\small{$\pm$2.7} & 55.6\small{$\pm$2.8} & 64.5 & 66.3 & 56.9\small{$\pm$2.5} & 60.6\small{$\pm$2.5} \\
			\hline
		\end{tabu}
	\end{center}
\end{table*}

\begin{table*}[!t]
	
	\begin{center}
		\caption{Standard domain adaption (left part) and label distribution shift (right part) based classification results for typical DA methods.}
		\label{tab5}
		\begin{tabu}{|c|c|c|[2pt]c|c|[2pt]c|c|[2pt]c|c|[2pt]c|c|[2pt]c|c|}
			\hline
			{} & \multicolumn{4}{|c|[2pt]}{VDA} & \multicolumn{4}{|c|[2pt]}{EasyTL} & \multicolumn{4}{|c|}{SP} \\ 
			\cline{1-13}
			{} & WC & WC$^{*}$ & WC & WWC$^{*}$ & WC & WC$^{*}$ & WC & WWC$^{*}$ & WC & WC$^{*}$ & WC & WWC$^{*}$ \\ 
			\hline\hline
			D$_1$ & 50.5 & 52.4 & 41.2\small{$\pm$7.2} & 43.9\small{$\pm$6.8} & 47.1 & 50.5 & 41.3\small{$\pm$6.5} & 46.1\small{$\pm$7.2} & 46.1 & 49.4 & 40.4\small{$\pm$7.0} & 45.8\small{$\pm$6.2} \\
			\hline
			D$_2$ & 82.8 & 83.6 & 75.6\small{$\pm$3.7} & 78.5\small{$\pm$3.6} & 83.5 & 84.7 & 74.8\small{$\pm$4.8} & 78.3\small{$\pm$4.2} & 85.1 & 86.7 & 78.4\small{$\pm$4.7} & 81.0\small{$\pm$4.7} \\
			\hline
			D$_3$ & 58.8 & 60.5 & 50.8\small{$\pm$2.5} & 54.1\small{$\pm$2.5} & 63.8 & 65.5 & 55.4\small{$\pm$2.5} & 59.0\small{$\pm$2.4} & 66.4 & 68.4 & 58.5\small{$\pm$2.8} & 62.7\small{$\pm$2.5} \\
			\hline
		\end{tabu}
	\end{center}
\end{table*}

\begin{figure}[t]
	\begin{center}
		\includegraphics[width=1.0\linewidth]{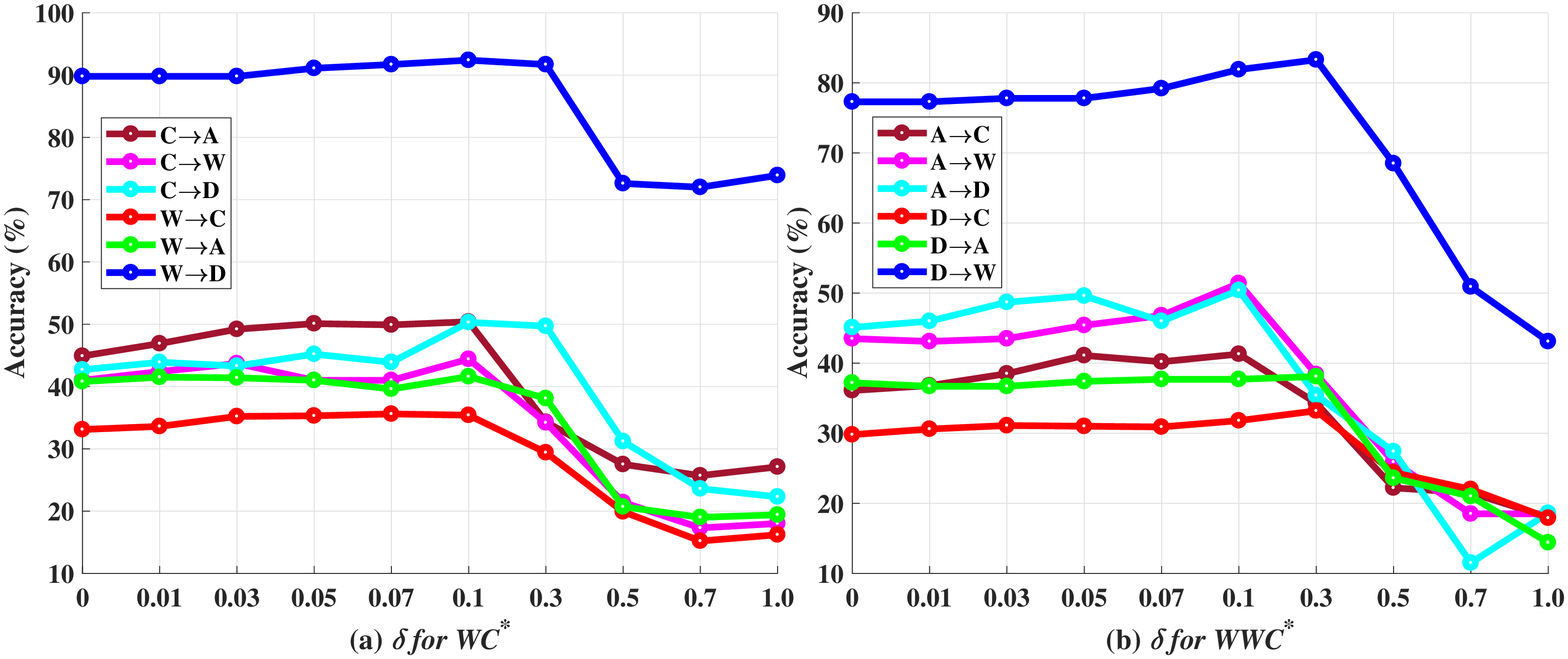}
	\end{center}
	
	\caption{Sensitivity analysis for the hyper-parameter $\delta$.}
	\label{fig3}
\end{figure}

\subsection{Ablation Study}
\label{Abation Study}
\par We further verify the effectiveness of WC$^*$, WWC by inspecting the feature distribution distance and HSI metric, and their feature visualizations could be found in Appendix $\textbf{\textit{C}}$.
We run Eq.~\eqref{eq21} with $\lambda\rightarrow +\infty$ (PCA), $\delta=0$ (WC), $\delta=0.1$ (WC$^*$), $\textbf{\textit{K}}_{yy}=\textbf{\textit{K}}_{yy}^4, \delta=0$ (WWC). Then we compute the aggregate feature distribution distance and HSI metric of each method on their induced embeddings. Note that, in order to compute the true distance or metric, we have to use the groundtruth labels instead of the pseudo ones. However, the groundtruth target labels are only used for verification, not for learning procedure, and we show them in  Fig.~\ref{fig2}.

\par From Fig.~\ref{fig2}, we can have these observations. 1) Without DA (PCA), the feature distribution distance is the largest, \textit{i.e.},  Fig.~\ref{fig2}(b). 2) With DA (WC), the HSI distances are reduced largely, and the proposed WC$^*$ can prompt feature-label correlation, \textit{i.e.}, Fig.~\ref{fig2}(a). 3) In label distribution shift, WC misaligns the feature distributions, and the designed label kernel (WWC) can realize more effective distribution matching \textit{i.e.}, Fig.~\ref{fig2}(b).

\subsection{Sensitivity of Hyper-Parameter $\delta$}
\label{sensitivity}
\par The proposed strategies for JMMD only entails one hyper-parameter $\delta$, and we conduct sensitivity analysis to validate that the optimal results could be achieved under a stable range. We only report target recognition results on 6 standard domain adaptation tasks (WC$^*$) and 6 label distribution shift scenario (WWC$^*$) from Office10-Caltech10 dataset, and similar trends on all other evaluations are proved, but not shown here due to space limitation.

\par We run the model Eq. \eqref{eq21} with WC$^*$ and WWC$^*$ with varying values of $\delta$. Then, we plot classification accuracy \textit{w.r.t.} different values of $\delta$ in Fig.~\ref{fig3}, and choose $\delta\in [0,1]$, which embodies the importance of the proposed strategy of feature-label correlation reinforcement. Moreover, it displays parameter's stability as the resultant classification accuracy %remains
often achieves its optimal value around 0.1.
\section{Conclusions}
% \vspace{-8pt}
The problem of JMMD is not fully explored in the field of DA due to its implementation difficulty. To overcome this problem, in this paper, we propose a unified JMMD framework that includes the previous popular marginal, class conditional and weighted class conditional distribution distances. In addition, we also prove that JMMD will lead to degradation of feature-label dependence that is beneficial to classification, and it is sensitive to the label distribution shift when the the label kernel is the weighted class conditional one. To remedy these issues, a HSI criterion based technique is taken to boost the feature-label dependence (a novel MMD matrix), and a new label kernel is designed to alleviate label distribution shift. Comprehensive tests carried on some benchmark datasets show promising performance of the unified JMMD method with those two strategies. Future work may lie in devising more desirable label kernels to address the problem of pseudo target labels in DA.

\section*{Acknowledgment}

This work is supported in part by the National Natural Science Foundation of China (NSFC) under Grants No.61772108, No.61932020, No.61976038, No.U1908210 and No.61976042.

\ifCLASSOPTIONcaptionsoff
  \newpage
\fi



%\bibliographystyle{IEEEtran}

\bibliographystyle{IEEEtran}
\bibliography{IEEETrans_our}

\iffalse
\begin{IEEEbiography}{Michael Shell}
Biography text here.
\end{IEEEbiography}

\begin{IEEEbiographynophoto}{John Doe}
Biography text here.
\end{IEEEbiographynophoto}

%\newpage

\begin{IEEEbiographynophoto}{Jane Doe}
Biography text here.
\end{IEEEbiographynophoto}
\fi

%\vfill


\end{document}